%% file: main.tex
\documentclass[10pt,twocolumn,letterpaper]{article}

\usepackage[pagenumbers]{cvpr} %
\usepackage{booktabs} %
\usepackage{multirow} %
\usepackage{enumitem}

\usepackage{academicons}
\usepackage{fontawesome}

\input{preamble}
\definecolor{cvprblue}{rgb}{0.21,0.49,0.74}
\usepackage[pagebackref,breaklinks,colorlinks,allcolors=cvprblue]{hyperref}

\def\paperID{63} %
\def\confName{EarthVision}
\def\confYear{2026}

\title{Pretrain Where? Investigating How Pretraining Data Diversity Impacts Geospatial Foundation Model Performance}

\author{
Amandeep Kaur$^{1}$\textsuperscript{ \faEnvelope} \quad Mirali Purohit$^{1*}$ \quad Gedeon Muhawenayo$^{1*}$ \\
Esther Rolf$^{2}$ \quad Hannah Kerner$^{1}$ \\
$^{1}$Arizona State University \quad $^{2}$University of Colorado Boulder \\
}

\begin{document}
\maketitle
\input{sec/0_abstract}    
\input{sec/1_intro}

\input{sec/2_related_work}
\input{sec/3_method}

\input{sec/4_results_n_discussion}
\input{sec/5_conclusion}

\newpage
{
    \small
    \bibliographystyle{ieeenat_fullname}
    \bibliography{main}
}

\input{sec/X_suppl}

\end{document}

%% file: preamble.tex
\usepackage[T1]{fontenc}

\usepackage{multirow}

%% file: sec/0_abstract.tex
\begin{abstract}
  New geospatial foundation models introduce a new model architecture and pretraining dataset, often sampled using different notions of data diversity. Performance differences are largely attributed to the model architecture or input modalities, while the role of the pretraining dataset is rarely studied. To address this research gap, we conducted a systematic study on how the geographic composition of pretraining data affects a model's downstream performance. We created global and per-continent pretraining datasets and evaluated them on global and per-continent downstream datasets. We found that the pretraining dataset from Europe outperformed global and continent-specific pretraining datasets on both global and local downstream evaluations. To investigate the factors influencing a pretraining dataset's downstream performance, we analysed 10 pretraining datasets using diversity across continents, biomes, landcover and spectral values. We found that only spectral diversity was strongly correlated with performance, while others were weakly correlated. This finding establishes a new dimension of diversity to be accounted for when creating a high-performing pretraining dataset. We open-sourced 7 new pretraining datasets, pretrained models, and our experimental framework at https://github.com/kerner-lab/pretrain-where.

\def\thefootnote{\faEnvelope}\footnotetext{ Corresponding Author: akaur64@asu.edu}\def\thefootnote{\english{footnote}}
\def\thefootnote{*}\footnotetext{ Equal Contribution}\def\thefootnote{\english{footnote}}

\end{abstract}

%% file: sec/1_intro.tex
\section{Introduction}
\label{sec:intro}

Pretraining large models on unlabeled data followed by task-specific fine-tuning has become a standard workflow in machine learning \cite{devlin-etal-2019-bert, dosovitskiy2021an, Howard2018UniversalLM}. Pretrained remote sensing/geospatial foundation models (RSFMs) have made significant progress in tasks like mapping landcover \cite{cong2022satmae}, crop type  \cite{Manas_2021_ICCV}, floods \cite{jakubik2023foundation}, and population estimation \cite{rolf2021generalizable}, domains where labelled data are scarce, expensive, and unevenly distributed across regions.

\begin{table}
  \caption{Overview of popular remote sensing foundation models (RSFMs) and the specific sampling focus used in their pretraining datasets.}
  \label{tab:model_datasets}
  \centering
  \begin{tabular}{@{}lc@{}}
    \toprule
      Model & Sampling focus \\
      \midrule
        SeCo~\cite{Manas_2021_ICCV} & Around city centers\\
        CROMA~\cite{fuller2023croma} & Around city centers\\
        Prithvi-V2~\cite{11296896} & Balanced landcover\\
        SeCo-Eco~\cite{Plekhanova_2025_CVPR} & Across uniform grid \\
        Galileo~\cite{pmlr-v267-tseng25a} & Balanced landcover \\
        MMEarth~\cite{nedungadi2024mmearth} & Balanced biomes \\
        SatlasNet~\cite{Bastani_2023_ICCV} & Biased to Global North \\
        MajorTOM~\cite{Francis2024MajorTE} &  Across uniform grid\\
        CopernicusFM~\cite{Wang_2025_ICCV} & Across uniform grid\\
    \bottomrule
  \end{tabular}
\end{table}

The widespread availability of unlabeled remote sensing data enables numerous spatial sampling strategies for creating large-scale pretraining datasets. In the absence of pretraining data standards, most studies proposing architectural contributions also come with new pretraining datasets. Table \ref{tab:model_datasets} shows that existing models rely on pretraining datasets with widely varying spatial spreads or distributions. 

Apart from SSL4Eco~\cite{Plekhanova_2025_CVPR}, prior works rarely isolate the contribution of the pretraining dataset to downstream performance, leaving the effects of their dataset distribution choices unclear. This is a significant gap: without accounting for the effects of pretraining data design choices, architectural progress is difficult to interpret and not necessarily additive.

\begin{figure}[t]
  \centering
    \includegraphics[width=0.9\linewidth]{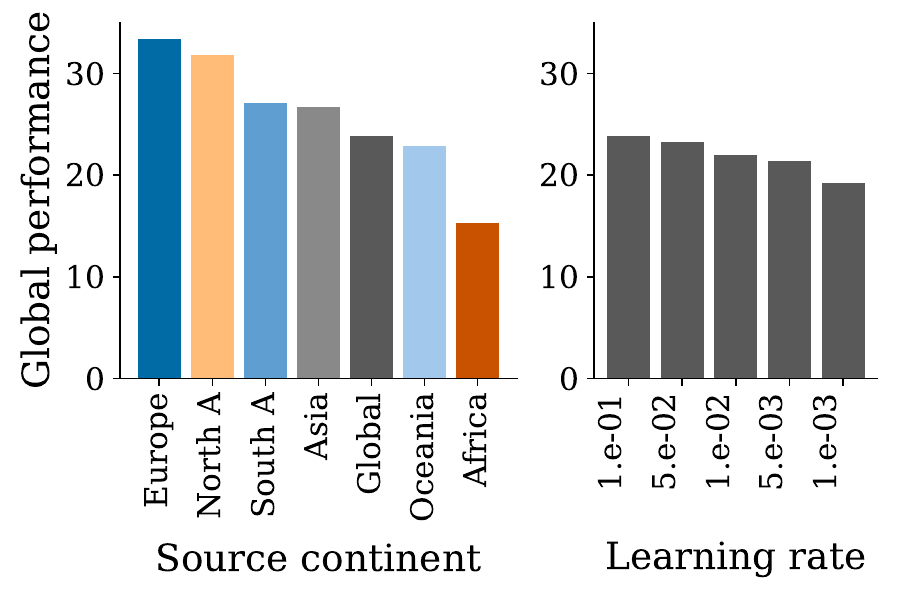}
  \caption{Downstream performance differences caused by varying the source continent of pretraining data at learning rate = 0.1 (left) vs. the learning rate with pretraining dataset = Global pretraining dataset (right) while holding everything else constant. The impact of the pretraining data's source continent is comparable to a major hyperparameter. North A = North America, South A = South America.}
  \label{fig:pre_vs_lr}
  \vspace{-1.8em}
\end{figure}

We studied the impact of varying the spatial distribution of pretraining data, treating continents as distinct spatial contexts. In Figure \ref{fig:pre_vs_lr} (left), we show that changing the source continent of pretraining data can have a surprisingly large impact on downstream performance. We compare this to the impact of varying a core hyperparameter, i.e., learning rate (right), and show that varying the source continent has a greater impact. 

\begin{figure*}
  \centering
    \centerline{\includegraphics[width=\linewidth]{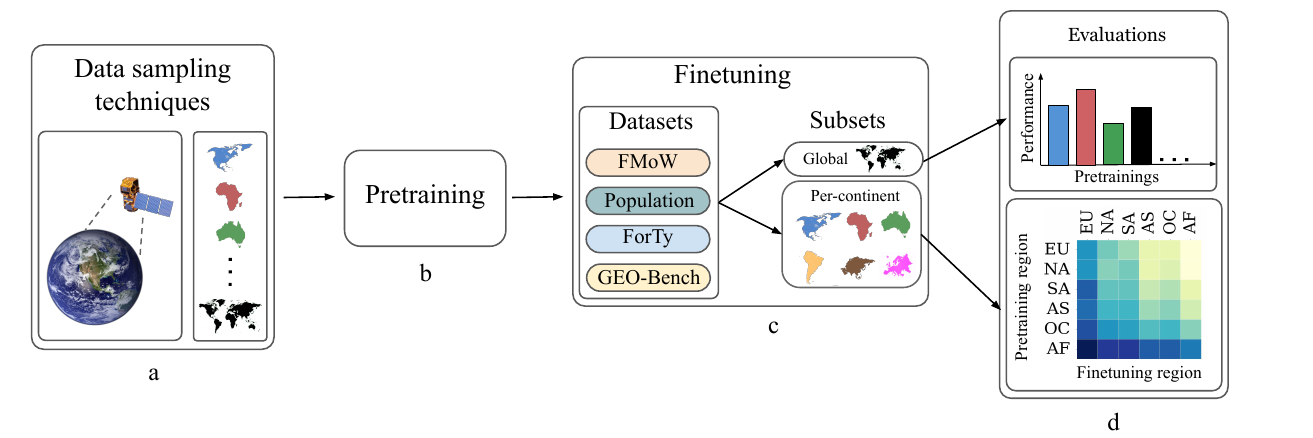}}
    \caption{
      Proposed pipeline to evaluate the performance of different pretraining datasets on a diverse set of downstream tasks. The procedure follows a) creation of spatially varying pretraining datasets, b) pretraining the model, c) finetuning on global and local subsets of downstream datasets, d) ranking and analysing pretraining datasets based on downstream performance.
    }
    \label{fig:hero}
\end{figure*}

While sweeping the learning rate is often feasible, sweeping over pretraining datasets is computationally expensive and therefore prohibitive. The search space grows even larger when considering alternative geographic units such as biomes or landcover types (Table \ref{tab:model_datasets}). 

To address this, we analysed the correlation between a pretraining dataset’s downstream performance and several measures of geographic data diversity, including diversity across continents, biomes and landcover types. The observed correlations hint towards a practical way to guide dataset selection without exhaustive experimentation. We list our \textbf{main contributions} as follows:
\begin{enumerate}
    \item We performed the \textbf{first systematic empirical study to understand the impact of the spatial distribution of pretraining data} on a model's downstream task performance.
    \item We reveal \textbf{surprising findings about the impact of pretraining data}. We show that matching the geographic composition (in terms of source continents) between pretraining and downstream data does not lead to optimal performance. We observe a consistent ranking among pretraining datasets, with Europe-only pretraining achieving the strongest results.

    \item We tested various measures of diversity in a pretraining dataset and \textbf{found that per-sample spectral entropy is strongly correlated with downstream performance}, while diversity across continents, biomes, and landcover types is weakly correlated. 
    \item We share \textbf{7 spatially varying pretraining datasets}, which can be downloaded and sampled for further experiments. Reproducibility is also supported by releasing \textbf{data splits and a model zoo}.
\end{enumerate}

%% file: sec/2_related_work.tex
\section{Related work}
\label{sec:related_work}

We review previous works in two key dimensions: (1) geospatial pretraining datasets and their spatial data distributions, and (2) studies analysing the influence of pretraining data distribution in both geospatial and natural image domains. 

\paragraph{Geospatial Pretraining Datasets.}
Early efforts like Functional Map of the World (FMoW)~\cite{Christie_2018_CVPR} and BigEarthNet~\cite{8900532} pioneered large-scale pretraining of RSFMs with supervised datasets. BigEarthNet, sized at ~600k samples, is limited to only Europe. FMoW, with ~1M samples, is a global 62-class scene classification dataset with a noticeable data collection bias towards the Global North. A similar bias is also observed in SatlasPretrain~\cite{Bastani_2023_ICCV}, with ~300M samples and 137 classes.

For large-scale self-supervised datasets, previous works have come up with various approaches for sampling the data in order to increase data diversity. SeCo \cite{Manas_2021_ICCV} sampled solely around the global city centres. SSL4EO-S12 \cite{10261879} followed the SeCo approach but also dropped overlapping patches. 

MMEarth \cite{nedungadi2024mmearth} assumed diversity in biomes and followed a balanced sampling scheme across 14 biomes. Prithvi-V2 \cite{11296896} and Galileo \cite{pmlr-v267-tseng25a} made landcover classes a central part of their sampling strategy. Galileo further ensured semantic diversity by running k-means clustering on the number of pixels per landcover class. 

Another popular approach used by SeCo-Eco \cite{Plekhanova_2025_CVPR}, MajorTOM \cite{Francis2024MajorTE}, and CopernicusFM \cite{Wang_2025_ICCV} to maximise diversity was to sample based on a uniform grid over the global landmass. MajorTOM defined rows and columns with a fixed on-ground distance of 10km between them. SeCo-Eco followed this approach while also ensuring a minimum distance of 23km between two points.  CopernicusFM first divided the globe into $\sim$1M 0.25° × 0.25° grid cells, followed by a Gaussian sampling around city centres.

Combining multiple pretraining datasets is another approach to increase diversity, as done in AnySat \cite{Astruc_2025_CVPR}, GeoPile \cite{Mendieta_2023_ICCV}, and Panopticon \cite{Waldmann_2025_CVPR}.

\paragraph{Impact of pretraining data distribution.}
The contribution of pretraining data distribution to a model's downstream performance is a widely studied area in the computer vision literature. To understand the robustness of CLIP, \cite{pmlr-v162-fang22a} systematically studied several possible factors, including pretraining dataset size, data distribution, language supervision, and learning objective, and found data distribution to be the biggest factor. 

\cite{ramanujan2023connection} analysed pretraining dataset size, image diversity, data sources (ImageNet vs iNaturalist), etc., to study how the properties of the pretraining dataset affected the robustness of a finetuned model. \cite{entezari2023the} performed extensive controlled experiments to prove that the choice of pretraining data distribution is essential for the few-shot transfer, but its role decreases as more data is made available for finetuning. \cite{longpre-etal-2024-pretrainers} analysed the impact of the temporal distribution of pretraining data, showing that models trained on temporally truncated datasets generalise poorly to post-cutoff samples. 

Within the geospatial community, \cite{purohit2025how} analysed the impact of different pretraining dataset sampling choices, including stratified by continent, stratified by biome, natural forest, and world cities, against uniform at random (UAR) sampling and zero-pretraining baselines. They concluded that balanced data compositions often outperformed region-specific ones. \cite{Plekhanova_2025_CVPR} isolated the impact of their pretraining dataset by retraining SeCo on their dataset and comparing it to the original model performance. They observed a substantial difference in the downstream performance attributable solely to the change in pretraining data. \cite{betti2025mapping} studied different definitions of geographic representativeness of training data for supervised machine learning with satellite data, but did not consider pretraining data.

In summary, prior works underscored that pretraining dataset distribution has a substantial impact on downstream model performance. But apart from \cite{purohit2025how}, controlled studies of pretraining data distributions and their influence on downstream performance remain largely unexplored.

%% file: sec/3_method.tex
\section{Method}
\label{sec:method}

Our experimental pipeline is shown in Figure \ref{fig:hero}. The following sections describe how we varied the pretraining data distribution (Section \ref{subsec:data_distribution}), downstream datasets (Section \ref{subsec:downstream_dataset}), the experimental details (Section \ref{subsec:experiment_design}), and diversity measures (Section \ref{subsec:diversity_measure}).

\subsection{Varying the data distribution}
\label{subsec:data_distribution}

To vary the pretraining data distribution, we followed the mathematical framework for dataset creation proposed by \citet{pmlr-v139-rolf21a}. They introduced partitioning of the data population into groups denoted by $g \in G$, where each sample $x_i$ is associated with a group $g_i$. The groups are required to be mutually exclusive and exhaustive with $\gamma_g = P_{(X, G)\sim D}[G = g]$ denoting the proportion of the population in group $g$, such that $\vec{\gamma} \in \Delta^{|G|}$.  For creating a dataset of size $n$, points are sampled from each group according to that group's allocation defined by 
\begin{equation}
\alpha_g := \frac{1}{n} \sum_{i=1}^{n} \mathbf{I}[g_i = g], \quad g \in G
\label{eq:group_allocation}
\end{equation}
where $\vec{\alpha} \in \Delta^{|G|}$. The group allocation vector $\vec{\alpha}$ governs the distribution of the final dataset created. 

In the context of the spatial distribution of geographic data, several relevant groupings exist, including continents, countries, biomes, ecoregions, etc. For this study, we worked with all continents, excluding Antarctica, due to its limited land use. So $\vec{\alpha} = [\alpha_{\text{Asia}}, \alpha_{\text{Africa}}, \alpha_{\text{Europe}}, \alpha_{\text{North America}}, \alpha_{\text{South America}}, \alpha_{\text{Oceania}}]$. We analysed a few extremes of the search space of $\vec{\alpha}$ defined as follows:
\begin{itemize}
    \item One-hot-$<$continent$>$: A fully biased $\vec{\alpha}$ with all samples coming from $<$continent$>$, e.g., for One-hot-Asia, the $\vec{\alpha}$ is $[1, 0, 0, 0, 0, 0]$.
    \item Global: A balanced $\vec{\alpha}$ with the same number of samples from each continent, i.e., $\vec{\alpha} = [1/6, 1/6, 1/6, 1/6, 1/6, 1/6]$.
    \item Zero-pretraining: A no-pretraining baseline where the model is initialised with random weights.
\end{itemize}

To control the dataset scale, we matched all pretraining datasets to the size of FMoW (700k samples), the original pretraining dataset of SatMAE \cite{cong2022satmae}.

\subsection{Downstream datasets}
\label{subsec:downstream_dataset}
We worked with global and per-continent subsets of large-scale global downstream tasks. We describe each global downstream task below. NOTE: For simplicity, we refer to FMoW-Sentinel as FMoW in the remainder of the paper.

\begin{itemize}
    \item \textbf{FMoW \cite{cong2022satmae}:} A large-scale urban scene classification dataset comprising 62 classes (e.g., amusement park, crop field, solar farm, swimming pool). 
    \item \textbf{MOSAIKS population density estimation:} A global regression task. We generated this dataset by exporting Sentinel-2 images using coordinates obtained from \citet{rolf2021generalizable}.
    \item \textbf{ForTy \cite{11243780}:} A large-scale landcover segmentation dataset with 8 classes: natural, planted, tree crops, other vegetation, built, water, ice, and bare.
    \item \textbf{GEO-Bench \cite{lacoste2023geobench}:} A global benchmark consisting of both classification and segmentation tasks, that vary in size and complexity. We worked with 6 tasks containing Sentinel-2 imagery: m-eurosat, m-bigearthnet, m-brick-kiln, m-so2sat, m-cashew-plantation, m-sa-crop-type. Individual task details can be found in the supplementary material.
\end{itemize}

\noindent\textbf{{Global and local subsets.}} The global subsets were constructed by sampling an equal number of samples from each continent. Per-continent subsets were constructed by drawing samples exclusively from a single continent. We denote global subsets by appending –global (e.g., FMoW-global) and per-continent subsets by appending the continent name (e.g., FMoW-Asia). All subsets are limited to fewer than 5k labelled samples. For balanced global and per-continent FMoW subsets, we used the 20 most frequent classes in each subset.

\subsection{Experimentation Details}
\label{subsec:experiment_design}
\paragraph{Model.}
We employed SatMAE \cite{cong2022satmae}, a Masked Autoencoder (MAE) based foundation model for remote sensing. We pretrained the ViT-Base variant using the authors’ released code and default parameters and hyperparameters. 

We chose SatMAE as our base model, as it is one of the most widely used foundation models in the geospatial domain. As SatMAE shares the vision transformer architecture and masked autoencoder learning objective with several geospatial foundational models, e.g., ScaleMAE \cite{reed2023scale}, CROMA \cite{fuller2023croma}, Prithvi \cite{jakubik2023foundation}, msGFM \cite{han2024bridging}, SatMAE++ \cite{noman2024rethinking}, MA3E \cite{li2024masked}; we expect our findings to translate to such RSFMs. 

\citet{entezari2023the, ramanujan2023connection} showed that the choice of pretraining data source is a major determinant of transfer performance across different architectures and learning objectives like supervised pretraining and CLIP \cite{pmlr-v139-radford21a}. This points to a possible generalisation of our results to an even broader set of RSFMs. Additional method details in Appendix A.

\paragraph{Finetuning details.}
Downstream performance is evaluated using linear probing, i.e., training a linear classification, regression, or segmentation head while keeping the pretrained weights frozen. We chose the best-performing learning rate from 
$\{1,3,5,8\}\times\{10^{-1},10^{-2},10^{-3},10^{-4},10^{-5}\}$ based on validation performance.

We also used kNN and full finetuning procedures to evaluate our pretraining datasets on the FMoW global task. For kNN, we swept over K values of 1, 3, 5, 10, 20, 40, 80, 160, and 320 and reported the test accuracy of the K that achieved the highest validation accuracy. For full fine-tuning, we swept learning rates of $\{1,3,5,8\}\times\{10^{-1},10^{-2},10^{-3},10^{-4}\}$.

For downstream evaluations, we use the author's released code when available and default parameters and hyperparameters, except for the learning rate. Additional model and downstream details in Appendix B and C.

\begin{table*}[ht]
  \caption{Performance comparison of pretrainings on global subsets across all downstream tasks. For GEO-Bench, \textbf{combination} denotes the aggregated result across 6 tasks. \textbf{Rankings} were calculated by averaging pretraining rankings corresponding to each downstream task.}
  \label{tab:global}
  \centering
  \begin{tabular}{@{}lccccc@{}}
    \toprule
    Pretraining &
    \multicolumn{1}{c}{FMoW-Sentinel} &
    \multicolumn{1}{c}{MOSAIKS Population} &
    \multicolumn{1}{c}{ForTy} &
    \multicolumn{1}{c}{GEO-Bench} &
    \multicolumn{1}{c}{Ranks} \\
     & (Accuracy $\uparrow$) & (R$^2$ Score $\uparrow$) & (F1 Score $\uparrow$) & (Combination $\uparrow$) & ($\downarrow$) \\
    \midrule
    One-hot-Europe & \textbf{0.33 $\pm$ 0.02} & \textbf{0.23 $\pm$ 0.02} & \textbf{0.37 $\pm$ 0.01} & \textbf{0.51} & \textbf{0.00} \\
    One-hot-North-America & 0.32 $\pm$ 0.02 & 0.22 $\pm$ 0.03 & 0.36 $\pm$ 0.02 & 0.49 & 1.00 \\
    One-hot-South-America & 0.27 $\pm$ 0.01 & 0.20 $\pm$ 0.03 & 0.36 $\pm$ 0.01 & 0.49 & 1.50\\
    One-hot-Asia & 0.27 $\pm$ 0.03 & 0.20 $\pm$ 0.03 & 0.35 $\pm$ 0.02 & 0.47 & 2.00 \\
    Global & 0.23 $\pm$ 0.02 & 0.17 $\pm$ 0.03 & 0.35 $\pm$ 0.01 & 0.46 & 2.75 \\
    One-hot-Oceania & 0.23 $\pm$ 0.02 & 0.14 $\pm$ 0.02 & 0.33 $\pm$ 0.02 & 0.45 & 3.50 \\
    One-hot-Africa & 0.15 $\pm$ 0.00 & 0.08 $\pm$ 0.03 & 0.31 $\pm$ 0.03 & 0.41 & 4.50 \\
    Zero pretraining & 0.12 $\pm$ 0.01 & 0.03 $\pm$ 0.04 & 0.30 $\pm$ 0.02 & 0.46 & 5.25 \\

    \bottomrule
  \end{tabular}
\end{table*}

\begin{table}[ht]
  \caption{Performance comparison of pretrainings on FMoW global subsets across kNN and full finetuning settings. The relative ordering of performance remained the same across all (kNN, linear probe, and full finetuning) settings.}
  \label{tab:knn_n_ff}
  \centering
  \begin{tabular}{lcc}
    \toprule
    Pretraining & kNN & Full finetuning \\
    \midrule
    One-hot-Europe & \textbf{0.31 $\pm$ 0.03} & \textbf{0.66 $\pm$ 0.03}\\
    One-hot-North-America & 0.26 $\pm$ 0.02 & 0.63 $\pm$ 0.02\\
    One-hot-South-America & 0.27 $\pm$ 0.03 & 0.62 $\pm$ 0.03\\
    One-hot-Asia & 0.27 $\pm$ 0.02 & 0.60 $\pm$ 0.02\\
    Global & 0.27 $\pm$ 0.02 & 0.54 $\pm$ 0.02\\
    One-hot-Oceania & 0.25 $\pm$ 0.01 & 0.46 $\pm$ 0.03\\
    One-hot-Africa  & 0.22 $\pm$ 0.00 & 0.30 $\pm$ 0.02\\
    Zero pretraining & 0.11 $\pm$ 0.01 & 0.22 $\pm$ 0.01\\

    \bottomrule
  \end{tabular}
\end{table}

\subsection{Diversity measures}
\label{subsec:diversity_measure}
Geographic diversity can be defined in multiple ways, but there is no consensus on which way is actually relevant to pretraining datasets. As seen in Table \ref{tab:model_datasets}, previous works employed various sampling approaches to maximise data diversity, e.g., SeCo focused on cities as centers of variation, and MMEarth ensured that data were captured across all biomes. 

We analysed the diversity of our pretraining datasets across various geographic features, including continents, biomes, and landcover types. For each geographic attribute, we defined the dataset diversity as the Shannon entropy, $H = -\sum_k p_k \log(p_k)$, of the dataset’s distribution over the corresponding classes.

\paragraph{Continent diversity.}
We quantified continent diversity by measuring the entropy of the distribution of images across continents. Let $\mathcal{I}$ denote the set of all images in the dataset, and let $\mathcal{C}$ denote the set of continents. Let $n_c$ denote the number of images originating from continent $c \in \mathcal{C}$, then $n_c = \sum_{i \in \mathcal{I}} \mathbf{1}[\text{continent}(i) = c]$ where $\mathbf{1}[\cdot]$ is the indicator function.

Let the total number of images in the dataset be $N = \sum_{c \in \mathcal{C}} n_c$. We defined the fractional distribution of images across continents as $p_c = \frac{n_c}{N}, \quad \forall c \in \mathcal{C}$, which defines a discrete probability distribution over continents.

We then defined continent diversity as the Shannon entropy of this distribution $H_{\text{continent}} = - \sum_{c \in \mathcal{C}} p_c \log p_c$. Higher entropy indicates that the dataset is evenly distributed across continents, whereas lower entropy indicates that it is concentrated on fewer continents.

\paragraph{Biome and Landcover diversity.}
We quantified the biome and landcover diversity of a dataset by measuring the entropy of its area distribution across biome and landcover classes. We used the RESOLVE biomes dataset~\cite{Dinerstein2017AnEA} with 15 biomes for biome mapping and the ESA WorldCover 10 m 2021 v200~\cite{zanaga2022esa} map with 11 landcover types for landcover mapping.

Let $\mathcal{I}$ denote the set of all images in the dataset, and let $\mathcal{C}$ denote the set of classes. When $\mathcal{C}$ represents biome classes, this yields the biome diversity. When $\mathcal{C}$ represents landcover classes, this yields the landcover diversity. 

Each image $i \in \mathcal{I}$ covers a spatial extent that may overlap multiple classes. Let $\mathbf{a}_i = \left[a_{i,c}\right]_{c \in \mathcal{C}}$
denote the area vector for image $i$, where $a_{i,c} \ge 0$ represents the spatial area of image $i$ belonging to class $c$. Then the total area covered by image $i$ is $A_i = \sum_{c \in \mathcal{C}} a_{i,c}$.

We aggregated these areas across the entire dataset to obtain the total area associated with each class, $A_c = \sum_{i \in \mathcal{I}} a_{i,c}, \quad \forall c \in \mathcal{C}$. Let the total spatial area of the dataset be $A_{\text{total}} = \sum_{c \in \mathcal{C}} A_c$.

We then computed the fractional area distribution across classes $p_c = \frac{A_c}{A_{\text{total}}}, \quad \forall c \in \mathcal{C}$, which defines a discrete probability distribution over classes. Finally, we defined the diversity as the Shannon entropy of this distribution, $H(\mathcal{C}) = - \sum_{c \in \mathcal{C}} p_c \log p_c$.

\paragraph{Spectral diversity.}
We defined spectral diversity using a sample-level entropy measure computed from the distribution of spectral values within each band for each image. Let $\mathcal{I}$ denote the set of all samples in the dataset, and let $\mathcal{B}$ denote the set of spectral bands. For a given sample $i \in \mathcal{I}$ and band $b \in \mathcal{B}$, let $\mathcal{X}_{i,b}$ denote the set of pixel values in band $b$.

We constructed a histogram for each band by partitioning the spectral value range into $K=100$ bins. Let $h_{i,b,k}$ denote the number of pixels in bin $k \in \{1, \dots, K\}$ for sample $i$ and band $b$. Let $N_{i,b} = \sum_{k=1}^{K} h_{i,b,k}$
denote the total number of pixels in band $b$ of sample $i$.

We defined the probability distribution over bins as $p_{i,b,k} = \frac{h_{i,b,k}}{N_{i,b}}, \quad k = 1, \dots, K$.
The spectral entropy for sample $i$ and band $b$ is then given by the Shannon entropy, $H_{i,b} = - \sum_{k=1}^{K} p_{i,b,k} \log p_{i,b,k}$.

We defined the sample-level spectral entropy as the average over all bands, $H_i = \frac{1}{|\mathcal{B}|}\sum_{b \in \mathcal{B}} H_{i,b}$. Finally, we defined the dataset-level spectral entropy as the mean spectral entropy across all samples, $H_{\text{spectral}} = \frac{1}{|\mathcal{I}|}\sum_{i \in \mathcal{I}} H_i$. 

Additional details for diversity anaylses in Appendix E.

\paragraph{Additional datasets.}
In addition to the seven pretraining datasets we created in this work, we extended the diversity analysis to include three published datasets that differ substantially in their sampling strategies: FMoW~\cite{cong2022satmae}, SSL4EO-S12~\cite{10261879}, and SSL4Eco~\cite{Plekhanova_2025_CVPR}.  
FMoW is the original pretraining dataset used by SatMAE. Although global, FMoW is a scene classification dataset, which means the samples are biased towards urban areas and infrastructure. Its sampling is also biased towards the Global North.
SSL4EO-S12 is a globally distributed unlabeled dataset sampled around city centers, following a strategy similar to SeCo~\cite{Manas_2021_ICCV}. Its sampling emphasised urban regions and areas of high human activity.
SSL4Eco adopted a uniform grid-based global sampling strategy, similar to MajorTOM~\cite{Francis2024MajorTE} and CopernicusFM~\cite{Wang_2025_ICCV}. SSL4Eco aimed to provide more spatially homogeneous coverage and encompass broader landcover coverage compared to SSL4EO-S12~\cite{Plekhanova_2025_CVPR}. \citet{Plekhanova_2025_CVPR} also observed that pretraining the SeCo model on SSL4Eco led to superior downstream performance compared to pretraining on SSL4EO-S12.

%% file: sec/4_results_n_discussion.tex
\section{Results}
\label{sec:results}

We present linear probing results from all global evaluation tasks in Table \ref{tab:global}, reporting mean performance and 95\% confidence intervals across five random data seeds. GEO-Bench performance was obtained by averaging results across its six constituent tasks. We also included an overall performance ranking, obtained by aggregating task-wise rankings based on mean performance across seeds. The best performance for each task is indicated in bold, and the best overall pretraining dataset is identified by the smallest rank. 

\paragraph{Performance of different pretraining strategies evaluated on global datasets.}
Table \ref{tab:global} shows that the choice of source continent alone leads to large performance differences (in the range of 10 to 21 metric points) across all global downstream tasks. We also note a consistent ranking of performance across all tasks, as shown by the rank column. All pretraining schemes outperformed Zero-pretraining across all tasks, except on GEO-Bench, where One-hot-Oceania and One-hot-Africa performed worse. 

One-hot-Europe achieved the best overall performance, with an aggregate rank of 0.00, outperforming Global pretraining, which has an aggregate rank of 2.75. One-hot-Europe improved over Global pretraining by 10\% on FMoW (33\% vs. 23\%), 0.06 R² points on MOSAIKS population density estimation (0.23 vs. 0.17), 0.02 F1 points on ForTy (0.37 vs. 0.35), and 0.05 points on the GEO-Bench aggregate score (0.51 vs. 0.46). These improvements are larger than the reported confidence intervals. One-hot-North-America, One-hot-South-America, and One-hot-Asia also outperformed Global pretraining.

The findings are further strengthened by the kNN and full finetuning evaluations on the FMoW global downstream task, as shown in Table~\ref{tab:knn_n_ff}. Similar to linear probing results, we saw a significant performance difference across the pretraining datasets, with an accuracy difference of 20\% and 44\% between the best and worst performing pretraining datasets for KNN and full finetuning. The relative ranking of the pretraining strategies also remained consistent.

\begin{figure}[t]
  \centering
    \includegraphics[width=1\linewidth]{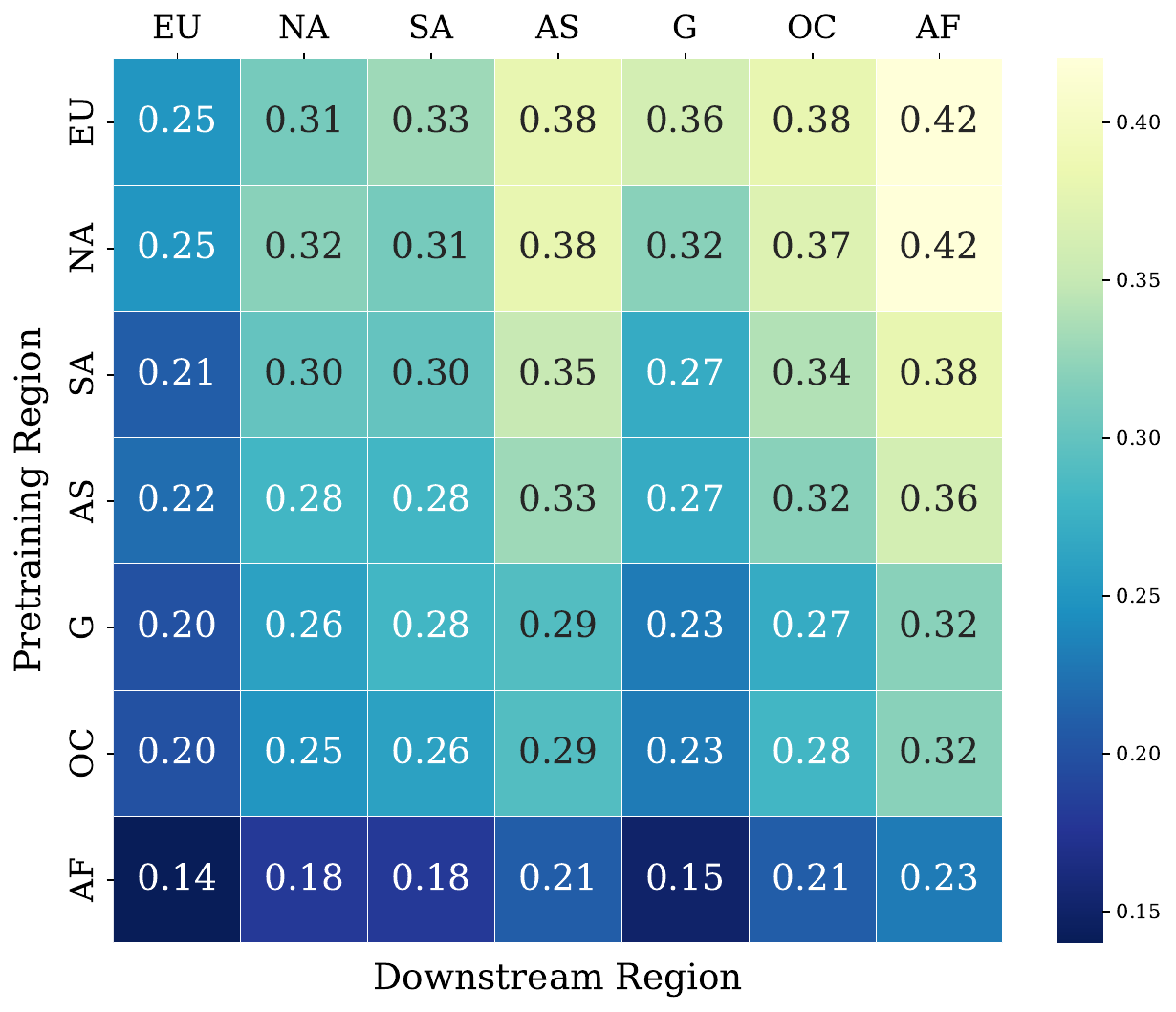}
  \caption{Performance comparison on FMoW subsets across pretraining data schemes. EU = Europe, NA = North America, SA = South America, AS = Asia, G = Global, OC = Oceania, AF = Africa.}
  \label{fig:fmow_sentinel}
\end{figure}

\paragraph{Performance of different pretraining strategies evaluated on continent-specific datasets.}
Figure \ref{fig:fmow_sentinel} reports performance on continent-specific subsets of the FMoW dataset. We note that, for each continent's downstream region, the performance ranking of pretraining schemes was the same as that observed on the Global downstream subset. Another trend seen across downstream regions was that models pretrained on any region achieved their highest performance on the FMoW-Africa subset and their lowest performance on the FMoW-Europe subset.  

We note that One-hot-Europe pretraining dataset outperformed One-hot-South-America, Global, and One-hot-Oceania on their respective downstream subsets. On the FMoW-North-America downstream subset, One-hot-North-America beat One-hot-Europe. Both performed the same on the FMoW-Asia and FMoW-Africa subsets. Apart from North America, no other pretraining outperformed One-hot-Europe on their respective downstream subsets. We got similar plots for the MOSAIKS population density and ForTy downstream tasks, which are available in the supplementary material. Additional local subset results in Appendix D.

Combining global and per-continent evaluations, we saw that pretraining strategies that performed well on global tasks also do well on continent-specific tasks, indicating consistent relative rankings in quality of pretraining datasets independent of the source continent of downstream data. 

\begin{figure}[t]
  \centering
    \includegraphics[width=\linewidth]{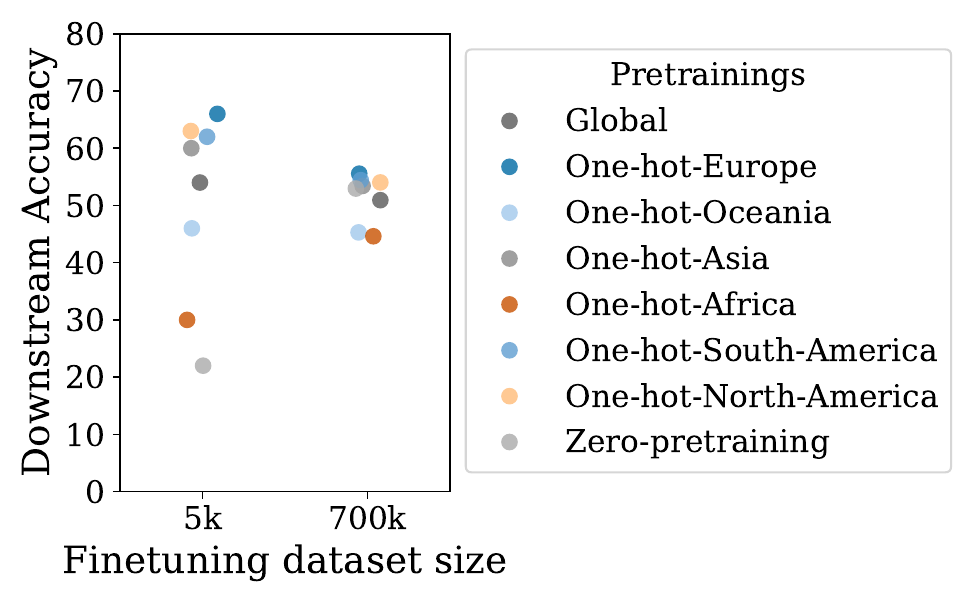}
    \label{fig:short-b}
  \caption{Performance comparison of our pretraining datasets when fully finetuned on the FMoW global 5k subset vs the 700k-sized FMoW original dataset. Performance variance reduced, but did not vanish.
}
  \label{fig:full_variance}
  \vskip -0.2in
\end{figure}

\paragraph{Impact of varying pretraining datasets on large-scale finetuning.}
\citet{purohit2025how, entezari2023the} have shown that the impact of pretraining reduces as the amount of finetuning data increases. We investigated whether this trend held in our case, as seen in Figure \ref{fig:full_variance}. To study the effect of larger finetuning datasets, we finetuned our pretrained models on the original 700k-sample global FMoW dataset. Given the substantially larger size of the finetuning dataset, we performed full finetuning rather than linear probing. For a fair comparison, we also conducted full finetuning on the 5k-sample global subset.

\begin{figure*}[ht]
  \centering
    \includegraphics[width=\linewidth]{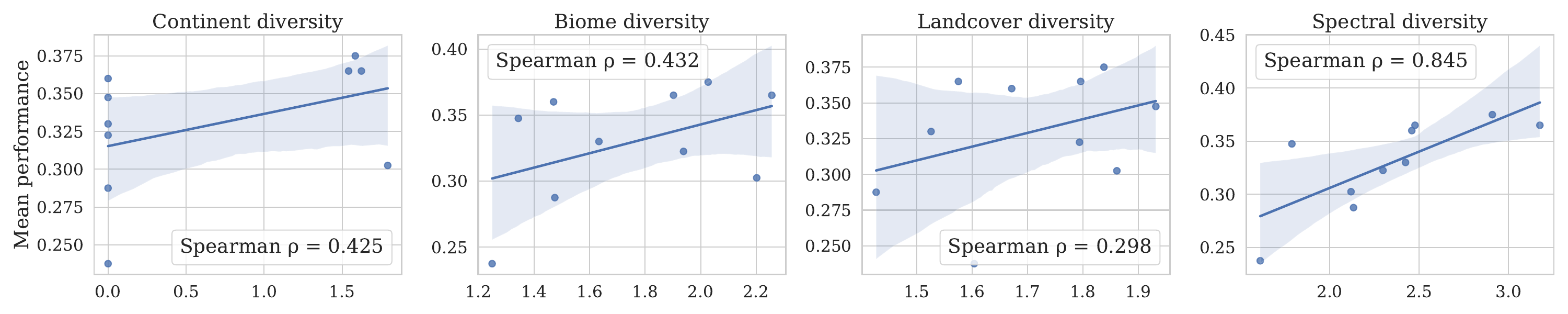}
  \caption{Correlation plots between mean model performance and diversity measures: continent, biomes, landcover and spectral diversity.}
  \label{fig:entropy_vs}
\end{figure*}

\paragraph{Correlation of downstream performance with diversity measures.}
We examined correlations between downstream performance and various measures of diversity, namely continent, biome, landcover, and spectral diversity. To calculate the correlation, we used the mean performance of pretrained models across four global downstream tasks: FMoW, Mosaiks population, ForTy, and combined GEO-Bench. We used the seven pretraining datasets we created, along with three others: FMoW, SSL4Eco, and SSL4Eo. 

Figure \ref{fig:entropy_vs} shows the correlations between all diversity measures and the mean downstream performance. We saw the correlations to be $\rho$ = 0.42, p = 0.221 for continent diversity, $\rho$ = 0.43, p = 0.213 for biome diversity, $\rho$ = 0.3, p = 0.403 for landcover diversity and $\rho$ = 0.84, p = 0.002 for spectral diversity. We note that the number of points is 10, so the correlations should be read with caution. 

\section{Discussion}
\label{sec:discussion}
We now discuss our results to understand the impact of pretraining data diversity on downstream performance. 

\paragraph{The Europe-only pretraining dataset consistently outperforms all other pretraining datasets on both global and local tasks.}
Our experimental results showed that the geographic composition of the pretraining dataset had a significant impact on downstream performance. As reported in Table~\ref{tab:global}, several single-continent pretraining schemes consistently outperformed the globally stratified pretraining dataset across all global downstream tasks. In particular, One-hot-Europe achieved a 10 percentage point improvement over Global pretraining on FMoW global downstream task. This finding contrasts with the general expectation that globally distributed pretraining data should yield superior performance on global downstream tasks compared to single-continent pretraining datasets.

Consistent performance trends across kNN, linear probing and full finetuning evaluation suggest that the observed differences in downstream performance are due to variations in pretraining data rather than artefacts of a particular evaluation setup.

For the continent-specific downstream subsets in Figure~\ref{fig:fmow_sentinel}, one might expect that pretraining on a given continent would yield optimal performance on its corresponding downstream subset due to geographic alignment and shared visual characteristics. Alternatively, globally distributed pretraining might be expected to generalise best across subset regions because of its more diverse learned features. However, neither pattern was observed. Instead, One-hot-Europe pretraining consistently achieved the strongest performance across nearly all continent-specific subsets.

Together, these results indicate that \textit{neither global distribution nor geographic alignment explains downstream performance in our setting; rather, the Europe-only pretrained model generalises most effectively across regions.}

\paragraph{The impact of pretraining persists even under extensive finetuning.}
We note that increasing the finetuning dataset size substantially reduced, but did not eliminate the performance differences caused by the choice of pretraining data. Even when the finetuning dataset was scaled to match the size of the pretraining data, a gap of more than 10 percentage points remained between the best- and worst-performing pretraining schemes (Figure \ref{fig:full_variance}). This indicates that finetuning on a large, task-specific dataset does not fully override differences in initialisation, and that \textit{the choice of pretraining dataset remains relevant even with large downstream datasets}.

\begin{figure}
  \centering
    \includegraphics[width=\linewidth]{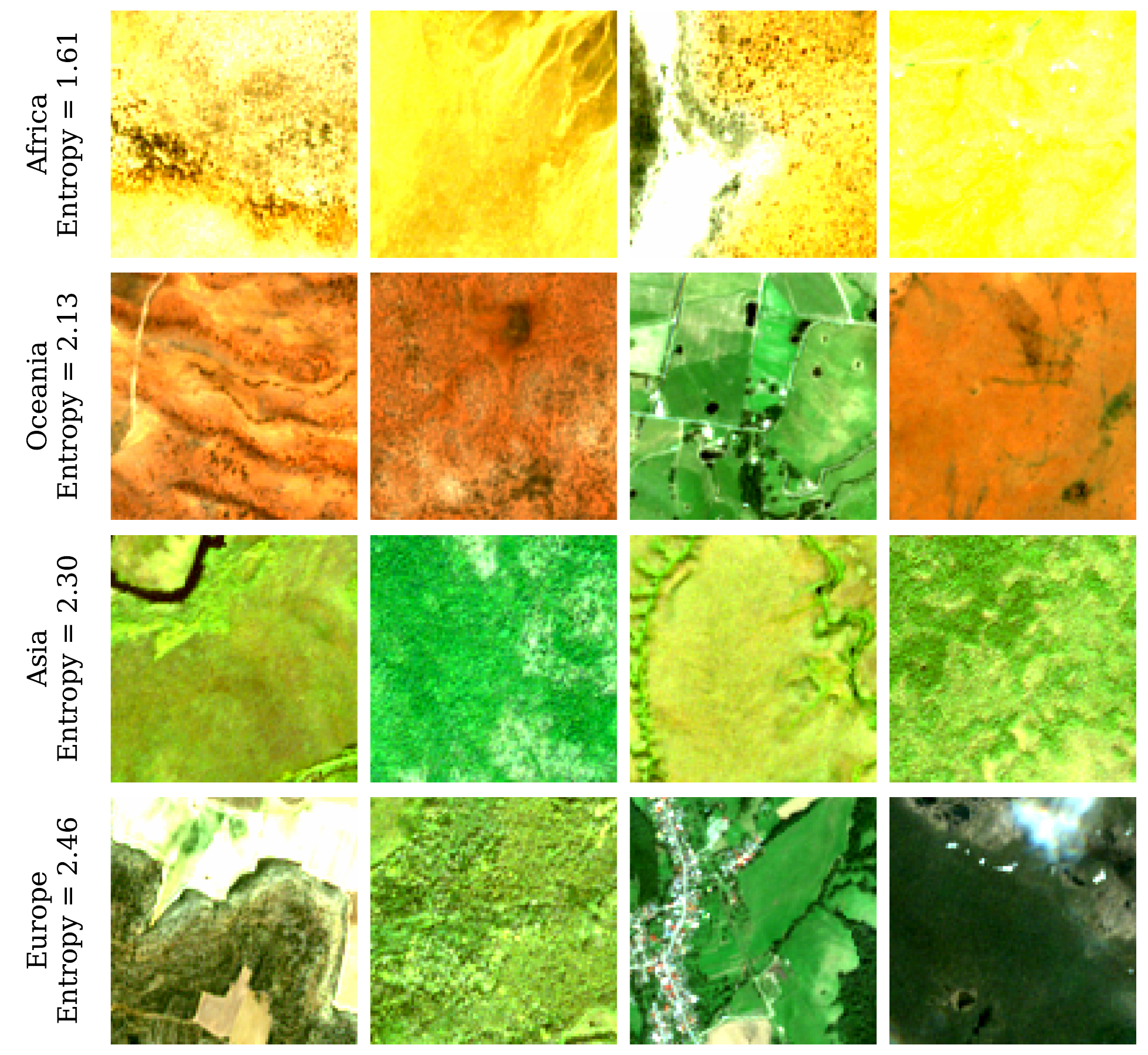}
  \caption{Rows show data samples from the One-hot-Africa, One-hot-Oceania, One-hot-Asia, and One-hot-Europe pretraining datasets. The mean spectral entropies of the datasets are 1.61, 2.13, 2.30, and 2.46, respectively.}
  \label{fig:samples}
  \vskip -0.2in
\end{figure}

\paragraph{A good pretraining dataset is geographically and spectrally diverse.}
Our findings raise a key question: if the geographic relevance of pretraining data (in terms of continent matching) does not correlate with downstream performance, then what does? What makes One-hot-Europe pretraining consistently outperform other pretraining datasets? 

Correlation analysis showed that commonly used dataset-level diversity measures such as diversity across continents, biomes, and landcover types were weakly associated with downstream performance. In contrast, sample-level spectral entropy exhibited a stronger correlation with performance ($\rho = 0.84$, $p = 0.002$). This suggests that differences in average sample complexity, rather than coarse measures of geographic coverage, are more likely driving the observed performance differences. 

Figure \ref{fig:samples} visualises representative samples from four pretraining datasets with increasing spectral entropy: One-hot-Africa (1.61), One-hot-Oceania (2.13), One-hot-Asia (2.30), and One-hot-Europe (2.46). The consistently strong performance of One-hot-Europe may be explained by its higher per-sample spectral entropy, indicating greater spectral complexity within individual training examples.

\section{Limitations and future work} 
Due to the computational cost of large-scale pretraining and evaluation, our study focused on a constrained but controlled experimental setting. \\
\\
\textbf{Models.} Several diverse geospatial foundation models exist (e.g., SatCLIP, CROMA, SeCo, Galileo), but we opted to experiment more thoroughly with a single architecture (SatMAE) due to the substantial computational requirements of pretraining across multiple datasets and evaluating on a diverse suite of downstream tasks. We expect the insights to translate to models with similar architectures, learning objectives, and scales. Extending this analysis to additional architectures is an important direction for future work.\\
\\
\textbf{Diversity as predictor of performance.} We evaluated the relationship between pretraining data diversity and downstream performance using ten pretraining datasets. While our results show a strong correlation between spectral diversity and downstream performance, evaluating a broader set of pretraining datasets would strengthen the generality of these findings.

%% file: sec/5_conclusion.tex
\section{Conclusion}
\label{sec:conclusion}

The role of the pretraining dataset remains relatively understudied for geospatial foundation models. In this work, we conducted the first systematic study isolating the effect of the geographic composition of pretraining data on downstream performance while controlling for model architecture and training procedure. 
We showed that, among six per-continent and one globally distributed pretraining datasets, Europe-only pretraining consistently outperformed both globally distributed and geographically aligned alternatives across global and local downstream tasks.

Importantly, we showed that downstream performance was strongly correlated with per-sample spectral entropy and only weakly correlated with continent, biome, and landcover diversity. This identifies spectral complexity as a key dimension of pretraining dataset design. Overall, our results underscore the need for principled dataset construction and provide empirical guidance for future geospatial foundation pretraining datasets.

\section{Acknowledgments}
\label{sec:acknowledgments}
This research was supported by funding from Google's Society-Centered AI Research program ("A Data-Centric Approach to Improve Geographic Equity in Geospatial ML", PI: Hannah Kerner). We acknowledge the Research Computing at Arizona State University for providing HPC resources that have contributed to the results reported in this paper. This work also used Bridges-2 at Pittsburgh Supercomputing Center through allocation cis240046p from the Advanced Cyberinfrastructure Coordination Ecosystem: Services \& Support (ACCESS) program, which is supported by National Science Foundation grants \#2138259, \#2138286, \#2138307, \#2137603, and \#2138296. We also thank Siddharth Sadhwani for his support throughout the development of this work, for helping maintain perspective during the process, and for offering valuable feedback on the paper’s narrative from an external viewpoint.

%% file: sec/X_suppl.tex
\clearpage
\setcounter{page}{1}
\setcounter{section}{0}
\maketitlesupplementary
\appendix

\section{Additional Experimental Details}
\subsection{Pretraining Dataset Construction}
For pretraining SatMAE, we created seven new pretraining datasets. We maintained a dataset size comparable to the original FMoW-Sentinel dataset (approx. 700k samples). The primary aspects in which we diverged are as follows:

\begin{enumerate}
    \item Scenes: FMoW-Sentinel is a 62-class scene classification dataset with images of urban structures. FMoW-Sentinel differs from the datasets we created as our images were chosen uniformly at random (UAR). Our pretraining captured random scenes without any structure.
    \item Image size: FMoW-Sentinel, being a scene classification dataset, has images of random sizes, with variable heights and widths ranging from 50 to 500 pixels. But the SatMAE dataset pipelines randomly scale and crop images to a final size of $96 \times 96$. To limit the dataset size, we restricted the downloaded image resolution to $96 \times 96$. Despite the reduced resolution, we retained random scaling and cropping.
\end{enumerate}

\subsection{Dataset Sampling Strategy}
We now provide details on the data sampling procedure for our pretraining datasets. We created seven datasets: six continent-specific ones and one global dataset. We sampled data using the open-source software QGIS, which performs sampling operations efficiently. For the continent-specific datasets, we loaded a world continent map and sampled the required number of points using the \textit{Random points in polygon} tool (Vector $>$ Research Tools). This sampled points uniformly at random, which satisfied the requirements of our use case.

\subsection{Dataset download}
We downloaded Sentinel-2 images from Microsoft Planetary Compute, which provides free access to the full Sentinel-2 dataset. We used Python multiprocessing to download images centered on the sampled latitudes and longitudes. We limited the maximum cloud cover to 20\% to obtain cloud-free samples. We initially sampled 900k points, retaining 700k for pretraining and 20k for validation to allow a buffer for download failures and missing images. We downloaded images from the year 2024. To further decrease dataset storage size, we normalised the images before saving, using the Sentinel-2 standard mean and standard deviation values from the SatMAE dataset loading pipeline available on GitHub. We made these datasets publicly available.

\subsection{Pretraining Hyperparameters}
We performed pretraining using the original dataset pipelines provided on SatMAE's GitHub. We also used the same hyperparameters. Since SatMAE's original hyperparameters list is a bit confusing, we provide the hyperparameters we used here, for a H100-80GB GPU:

\begin{itemize}[noitemsep]
    \item batch size = 256
    \item accum iter = 16
    \item blr = 0.0001
    \item epochs = 50
    \item warmup epochs = 5
    \item input size = 96 
    \item patch size = 8
    \item mask ratio = 0.75
    \item model = 'mae\_vit\_base\_patch16'
    \item model type = 'group\_c'
\end{itemize}

We pretrained for 50 epochs, as done for the original SatMAE pretraining, and noted convergence.

\subsection{Compute Infrastructure and Training Cost}
All experiments were run on resources on ASU Sol supercomputer \cite{HPC:ASU23} and PSC bridges-2 \cite{brown2021bridges2} resources. We utilized two distinct compute resources: NVIDIA A100-40GB (ASU Sol) and NVIDIA H100-80GB GPUs (PSC Bridges-2). We distributed the workload for pretraining and downstream tasks across both environments. The estimated runtimes for each setup are detailed below.

\paragraph{NVIDIA A100 (40GB) Benchmarks}

\begin{itemize}[noitemsep]
    \item Pretraining: $\approx 48$ hours
    \item FMoW Evaluation: $\approx 60$ minutes
    \item Mosaiks Evaluation: $\approx 20$ minutes
    \item ForTy Evaluation: $\approx 20$ minutes
    \item GeoBench Evaluation: $\approx 300$ minutes
\end{itemize}
\noindent \textbf{Total A100 Cost:} Pretraining ($48\text{h} \times 10$) + Downstream ($400\text{min} \times 11 \times 7 \times 5$).

\paragraph{NVIDIA H100 (80GB) Benchmarks}
\begin{itemize}[noitemsep]
    \item Pretraining: $\approx 10$ hours
    \item FMoW Evaluation: $\approx 15$ minutes
    \item Mosaiks Evaluation: $\approx 5$ minutes
    \item ForTy Evaluation: $\approx 5$ minutes
    \item GeoBench Evaluation: $\approx 70$ minutes
\end{itemize}
\noindent \textbf{Total H100 Cost:} Pretraining ($10\text{h} \times 10$) + Downstream ($100\text{min} \times 11 \times 7 \times 5$).

\noindent Since we utilised both server types, our true computational cost is split between the two. 

\section{Model and Training Details}

\subsection{SatMAE Architecture choices}
We utilised the ViT-B architecture for our experiments to optimise computational resources. As demonstrated in the SatMAE paper, the performance gap between ViT-B and ViT-L was minimal. We observed similar trends in our preliminary experiments; therefore, we proceeded with ViT-B to reduce training costs. We used the standard architecture without any modifications.

\subsection{Pretraining Objective and Optimisation}
We provided all the pretraining hyperparameters in A.4. We did not make any other changes apart from moving data normalisation to the download stage for faster dataset loading. The pretraining objective remained unchanged.

\subsection{Linear Probing Protocol}
We added linear classifiers for each downstream task, as the original SatMAE code was limited to full finetuning for classification tasks. For all downstream tasks, we fixed the training duration to 50 epochs, as we observed convergence within this timeframe. We used a cosine decay schedule with 5 warmup epochs and a batch size of 512. We swept over a large set of learning rates specific to each downstream task; we ensured the sweep is exhaustive by verifying that the optimal learning rate is bounded by lower-performing rates on both sides. To satisfy this criterion, we swept across a range of $\{1,3,5,8\}\times\{10^{-1},10^{-2},10^{-3},10^{-4},10^{-5}\}$ to identify optimal values.

\section{Downstream Datasets and Evaluation}
\subsection{FMoW-Sentinel Dataset Details and Splits}
We used the publicly available FMoW-Sentinel dataset. To create continent-specific subsets, we split the global dataset using QGIS. We imported the world continents map and the FMoW-Sentinel image polygons, then computed the intersection of the two vector layers. To create smaller subsets capped at $\approx 5$k samples, we restricted the dataset to the 20 most frequent classes per continent. To achieve a 70:15:15 (train:val:test) split, we limited sampling to $\approx 250$ images per class (175 for train, 38 for validation, and 38 for test). This resulted in a total dataset size of $20 \times 251 \approx 5\text{k}$ samples.

For the global subset, we sampled from the continent-specific subsets. From each subset, we selected 44 samples per class: 30 for training, 7 for validation, and 7 for testing. 

The total dataset size was $(30+7+7) \times 20 \times 6 \approx 5\text{k}$ samples. 

To generate 5 data seeds, we created 5 distinct versions of all subsets by changing the random seed during the sampling procedure.

\subsection{MOSAIKS Population Density Dataset}
We constructed this particular dataset. We took the labels with their latitude and longitude coordinates from the MOSAIKS paper. We downloaded the corresponding Sentinel-2 images similar to how we did for our custom pretraining datasets mentioned in A.3. Similar to FMoW-Sentinel, we divided the full dataset into per-continent partitions and then sampled ~5k points. We divided the points into 70:15:15-sized splits. 5 data seeds were created as done for FMoW-Sentinel.

\subsection{ForTy Segmentation Dataset}
We used the publicly available ForTy dataset in tfrecords format. We extracted point coordinates from the tfrecords and followed procedures similar to MOSAIKS for creating subsets.

\subsection{GEO-Bench}
\label{app:geobench}
We did not alter anything in the GEO-Bench datasets. We used the 6 datasets that contained data from Sentinel-2 and used the 1.00x partitions, which have 100\% data. We did not work with the data seeds given for GEO-Bench. We now give details for individual GEO-Bench tasks that we worked on:
    \begin{itemize}%
        \item \textbf{m-eurosat:} A 4k-sized, 10-class scene classification dataset with samples from Europe only.
        \item \textbf{m-bigearthnet:} A 22k-sized, 43-class land cover classification dataset with samples from Europe only.
        \item \textbf{m-brick-kiln:} A 17k-sized, 2-class brick kiln classification dataset with samples from Bangladesh, Asia only.
        \item \textbf{m-so2sat:} A 21k-sized, 17-class land cover classification with global samples.
        \item \textbf{m-cashew-plantation:} A 1.8k-sized, 7-class cashew plantation identification segmentation task with samples from Benin, Africa.
        \item \textbf{m-sa-crop-type:} A 5k-sized, 10-class crop type classification dataset with samples from South Africa, Africa.
    \end{itemize}

\subsection{Evaluation Metrics}
We used Accuracy, R2 score, and F1Score for FMoW-Sentinel, MOSAIKS, and Forty, respectively. For GEO-Bench, we worked with the standard metrics that came with each task, i.e., Jaccard for m-SA-crop-type and m-cashew-plantation, F1Score for m-bigearthnet, and Accuracy for the rest.

\section{Additional Results}

\subsection{Per-Continent Downstream Results}
We present per-continent downstream results drawn as heatmaps for MOSAIKS and ForTy tasks here:

\begin{figure}[htbp]
  \centering
    \includegraphics[width=\linewidth]{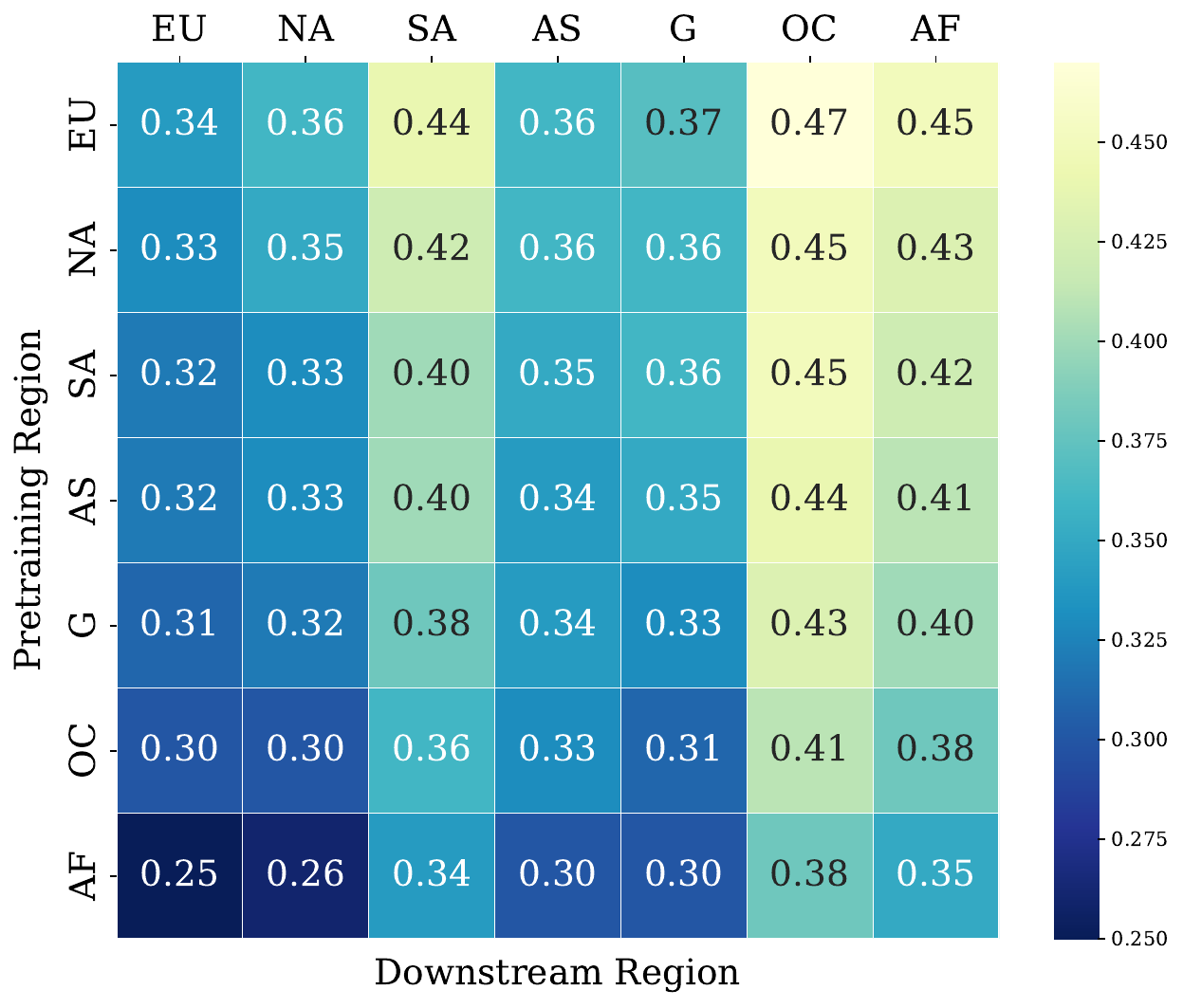}
  \caption{Performance comparison on ForTy global subsets across pretraining data schemes. In-distribution sampling is not always optimal.
}
  \label{fig:forty_heat}
      \vskip -0.2in

\end{figure}

\begin{figure}[htbp]
  \centering
    \includegraphics[width=\linewidth]{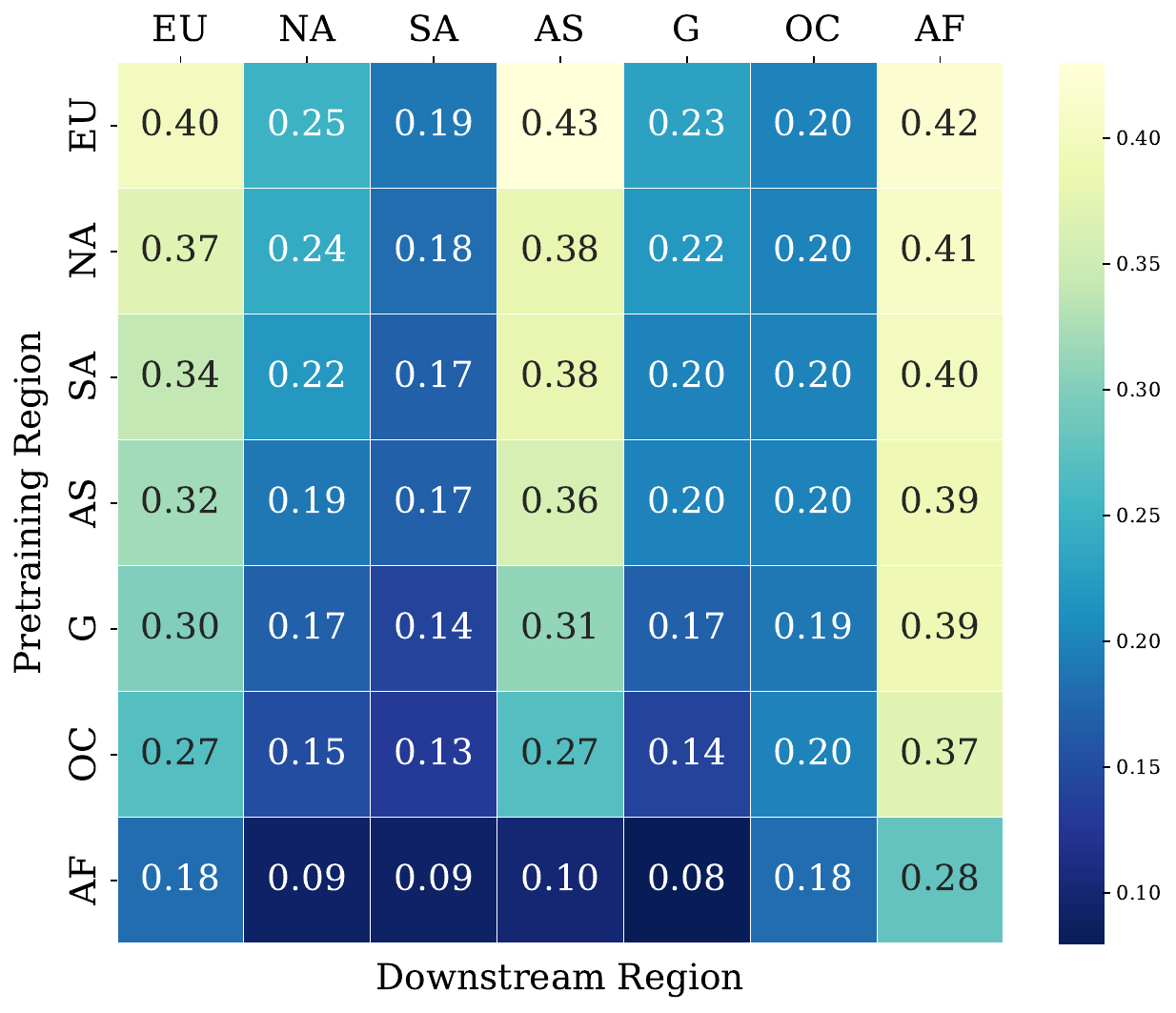}
  \caption{Performance comparison on MOSAIKS population density global subsets across pretraining data schemes. In-distribution sampling is not always optimal.
}
  \label{fig:mosaiks_heat}
      \vskip -0.2in

\end{figure}

\section{Dataset Diversity Analysis}

\subsection{Other pretrainings}
For the diversity analyses, we also worked with 3 additional published datasets, FMoW-Sentinel, SSL4Eo and SSL4Eco.
\begin{itemize}
    \item FMoW-Sentinel: FMoW-Sentinel is a 700k-sized 62-class scene classification dataset, originally used for unsupervised pretraining by SatMAE. The distribution is biased towards the Global North. The spatial distribution of FMoW is [0.21, 0.09, 0.35, 0.23, 0.08, 0.02] in terms of sample distribution for continents Asia, Africa, Europe, North America, South America, and Oceania.
    \item SSL4Eco: is a 1 M-sized global pretraining dataset. SSL4Eco is seasonal, i.e., captures 4 images from 4 seasons for 250k locations to reach 1M. The authors of SSL4Eco also distinguish SSL4Eco as a dataset with samples from all Copernicus landcover classes. For sampling, the dataset followed a uniform grid strategy from Major-TOM, but with 23km spacing between any two points. The spatial distribution is [0.32, 0.21, 0.07, 0.17, 0.12, 0.05]
    \item SSL4Eo: SSL4Eo is also sized at 1M, with sampling focused around city centers. The SSL4Eo authors chose city centers and then sampled around them in a radius of 50km using Gaussian sampling. They also ensured removal of overlapping samples. 
\end{itemize}

\subsection{Geographic Class Definitions (Continents, Biomes, Landcover)}
For diversity calculations, we use three approaches - continent-based, biome-based, and landcover-based. Notice that in each of these approaches, the dataset can be partitioned into multiple groups/classes. Specifically:
\begin{itemize}
    \item Continents: have 6 groups, namely Asia, Africa, Europe, North America, South America, Oceania.
    \item Biomes: According to the RESOLVE biome map, there are 15 biomes namely (1) Deserts \& Xeric Shrublands, (2) Tropical \& Subtropical Grasslands, Savannas \& Shrublands, (3) Boreal Forests/Taiga, (4) Tundra, (5) Tropical \& Subtropical Moist Broadleaf Forests, (6) Temperate Broadleaf \& Mixed Forests, (7) Temperate Grasslands, Savannas \& Shrublands, (8) Mediterranean Forests, Woodlands \& Scrub, (9) Montane Grasslands \& Shrublands, (10) Tropical \& Subtropical Dry Broadleaf Forests, (11) Temperate Conifer Forests, (12) Flooded Grasslands \& Savannas, (13) Tropical \& Subtropical Coniferous Forests, (14) Mangroves, (15)rock and ice.
    \item Landcover: According to ESA Worldcover 2021 v200, landcover has 11 classes namely (1) Tree cover, (2) Shrubland, (3) Grassland, (4) Cropland, (5) Built-up, (6)Bare / sparse vegetation, (7) Snow and ice, (8) Permanent water bodies, (9) Herbaceous wetland, (10) Mangroves, (11) Moss and lichen
\end{itemize}

\begin{figure}
  \centering
    \includegraphics[width=\linewidth]{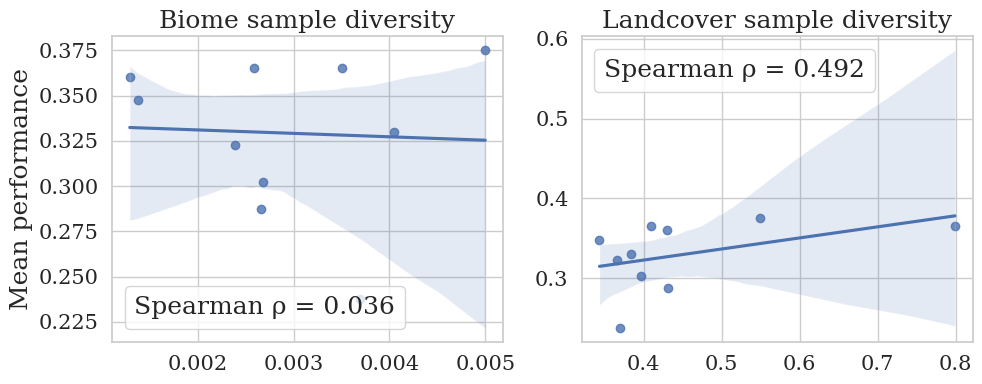}
  \caption{Correlation plots between mean model performance and diversity measures: sample biome and sample landcover diversity.}
  \label{fig:sample_diversity}
\end{figure}

\subsection{More diversity measures}
We now look at biome and landcover diversity from the sample entropy perspective similar to how the spectral diversity is calculated. We calculate entropy for each sample and then take the mean over the dataset. We found that sample biome diversity (shown in Figure  \ref{fig:sample_diversity}, left) had a lower correlation with downstream performance as compared to biome diversity. On the other hand, sample landcover diversity (shown in Figure \ref{fig:sample_diversity}, right) has a higher correlation than lancover diversity. 

Note that the definitions of biome and landcover diversity used in the main paper align more with biome/landcover based stratified sampling or sampling to ensure data from all biomes/landcovers for example, MMEarth \cite{nedungadi2024mmearth}, SSL4Eco \cite{Plekhanova_2025_CVPR}, Presto \cite{tseng2023lightweight}, etc. While the definitions defined here are related to diversity ensured within a sample, which is implicitely done in sampling approach of Galileo \cite{pmlr-v267-tseng25a}.  

\subsection{Per-band spectral diversity}
Since spectral diversity is calculated by first calculating entropy for a sample's band and then averaging over bands, we also find a dataset's diversity in terms of its individual bands. For this, we calculate a list of entropies for each sample corresponding to the bands, and then average them across the dataset. We show the values for the One-hot-Europe dataset in Table \ref{tab:spectral_bands}

\begin{table}
  \caption{Per-band spectral diversity for One-hot-Europe pretraining dataset.}
  \label{tab:spectral_bands}
  \centering
  \begin{tabular}{lcc}
    \toprule
    Band & Spectral entropy \\
    \midrule
    B2 (blue) & 2.20 \\
    B3 (green) & 2.38 \\
    B4 (red) & 2.22 \\
    B5 (red edge) & 2.46 \\
    B6 & 2.60 \\
    B7 & 2.63 \\
    B8 (NIR) & 2.56 \\
    B8A & 2.64\\
    B11 (SWIR 1) & 2.54 \\
    B12 (SWIR 2) & 2.37 \\
    \bottomrule
  \end{tabular}
\end{table}

We note that the RGB bands B4, B3 and B2 show the least entropy whereas bands like B6, B7, B8, B8A (Red Edge bands normally used for vegetation related tasks) have higher entropy.